\documentclass{article}

\usepackage[english]{babel}
\usepackage{microtype}
\usepackage{graphicx}
\usepackage{booktabs} 
\usepackage{multirow}
\usepackage{array}

\usepackage{hyperref}

\usepackage[accepted]{icml2025}

\usepackage{amsmath}
\usepackage{amssymb}
\usepackage{mathtools}
\usepackage{amsthm}

\usepackage[capitalize,noabbrev]{cleveref}

\theoremstyle{plain}

\theoremstyle{definition}

\theoremstyle{remark}

\usepackage[textsize=tiny]{todonotes}
\usepackage{subfig}
\newcolumntype{C}[1]{>{\centering\arraybackslash}p{#1}}
\icmltitlerunning{Submission and Formatting Instructions for ICML 2025}

\begin{document}

\twocolumn[
\icmltitle{Spec-LLaVA: Accelerating Vision-Language Models \\ with Dynamic Tree-Based Speculative Decoding}




\icmlsetsymbol{equal}{*}

\begin{icmlauthorlist}
\icmlauthor{Mingxiao Huo}{cmu,equal}
\icmlauthor{Jiayi Zhang}{unuk,equal}
\icmlauthor{Hewei Wang}{cmu,equal}
\icmlauthor{Jinfeng Xu}{hku}
\icmlauthor{Zheyu Chen}{polyu}
\icmlauthor{Huilin Tai}{cu}
\icmlauthor{Yijun Chen}{ucb}
\end{icmlauthorlist}

\icmlaffiliation{cmu}{Carnegie Mellon University}
\icmlaffiliation{unuk}{University of Nottingham}
\icmlaffiliation{hku}{The University of Hong Kong}
\icmlaffiliation{polyu}{The Hong Kong Polytechnic University}
\icmlaffiliation{cu}{Columbia University}
\icmlaffiliation{ucb}{University of California, Berkeley}


\icmlkeywords{Machine Learning, ICML}

\vskip 0.3in
]



\printAffiliationsAndNotice{\icmlEqualContribution} 

\begin{abstract}
Vision-Language Models (VLMs) enable powerful multimodal reasoning but suffer from slow autoregressive inference, limiting their deployment in real-time applications. We introduce Spec-LLaVA, a system that applies speculative decoding to accelerate VLMs without sacrificing output quality. Spec-LLaVA pairs a lightweight draft VLM with a large target model: the draft speculates future tokens, which the target verifies in parallel, allowing multiple tokens to be generated per step. To maximize efficiency, we design a dynamic tree-based verification algorithm that adaptively expands and prunes speculative branches using draft model confidence. On MS COCO out-of-domain images, Spec-LLaVA achieves up to 3.28$\times$ faster decoding on LLaVA-1.5 (7B, 13B) with no loss in generation quality. This work presents a lossless acceleration framework for VLMs using dynamic tree-structured speculative decoding, opening a path toward practical real-time multimodal assistants. Importantly, the lightweight draft model design makes the framework amenable to resource-constrained or on-device deployment settings. 
\end{abstract}

\section{Introduction}

Large vision-language models (VLMs), such as LLaVA~\cite{liu2023visual}, combine image understanding with language generation to enable rich multimodal interactions. However, their autoregressive decoding and large parameter sizes make inference slow. Generating a single response may require hundreds of forward passes through a 7B or 13B model, resulting in high latency that hinders real-time deployment~\cite{Huo2023HumanorientedRL, zhu2023fanuc}. Existing acceleration methods—such as quantization~\cite{frantar-gptq}, early exit~\cite{schuster2021cats}, or distillation~\cite{hinton2015distilling}—offer limited speedups (up to $\sim$3.5$\times$) and often degrade output quality or require extensive tuning. 

Recently, \emph{speculative decoding}~\cite{leviathan2023fast, chen2023accelerating} has emerged as a promising approach to accelerate language model inference without altering outputs. A lightweight draft model predicts tokens ahead, which the target model verifies in parallel. If matched, multiple tokens are accepted in one step, yielding lossless acceleration—identical outputs with lower latency. While prior works such as SpecInfer~\cite{miao2023specinfer}, EAGLE-2~\cite{li2024eagle}, OPT-Tree~\cite{wang2024opt}, and Sequoia~\cite{chen2024sequoia} apply this technique to LLMs, they focus on static sequences or predefined trees. For example, Wen et al.~\cite{wen2024ctc} proposed a CTC-based method to enhance acceptance in text-only decoding. Yet, speculative decoding remains unexplored for multimodal models, where higher output variability demands more flexible strategies~\cite{Lin2024JointPT}. Recent advances in multimodal recommendation also demonstrate that multi-level self-supervised objectives can effectively enhance cross-modal alignment, which can inform future draft–target calibration strategies \cite{xu2025mentor}. In parallel, kernelized optimal transport—most notably ITD and its distributed testing variants—offer tools to quantify uncertainty for draft–target calibration under noisy supervision~\cite{lin2023integrated, lin2024integrated, lin2025leveraging}. In particular, Lin and Ruszczynski~\cite{lin2023fast} introduce a fast dual subgradient optimization framework for integrated transportation distance, which can further enhance draft–target calibration efficiency. At the system level, federated computation of free-support transportation barycenters enables privacy-preserving distribution alignment across clients, which can complement noise-aware acceptance and branching policies~\cite{lin2025federated}. Meanwhile, the effect of pre-training data quality on acceleration remains underexplored; recent studies show that noisy supervision impacts convergence and robustness~\cite{10934976, chen2025impactnoisysupervisionfoundation}. Additionally, recent advances in sparse-view 4D reconstruction, such as MonoFusion~\cite{wang2025monofusionsparseview4dreconstruction}, highlight the potential of integrating high-fidelity visual information to enhance multimodal speculative decoding strategies.

We present Spec-LLaVA, a system that extends speculative decoding to VLMs. It combines a small, distilled draft model (68M or 160M parameters) with a full-scale LLaVA-1.5 target. Both draft and target take image and text inputs, enabling speculative token trees guided by visual grounding. This grounding imposes semantic constraints that improve alignment between draft and target distributions. The compact draft model also enables low-latency inference in resource-constrained settings such as mobile or edge devices~\cite{xu2024ondevice}, where full VLMs are impractical.

To maximize accepted tokens, we introduce a dynamic tree-based verification algorithm inspired by OPT-Tree~\cite{wang2024opt} and adapted for uncertainty-aware decoding. When confident, the draft expands a narrow tree; when uncertain, it explores multiple branches. A leaf-to-root verification strategy ensures exact match with the target model, enabling lossless acceleration. This architecture supports a hybrid inference setup, where speculative generation occurs locally, with periodic verification deferred to a larger model in the cloud or server. Our contributions are as follows:

\begin{itemize}\setlength{\itemsep}{1pt}
    \item We propose Spec-LLaVA, a speculative decoding method for vision-language models, achieving lossless acceleration without compromising output quality.
    \item We develop a dynamic tree-based verification algorithm for Spec-LLaVA that adapts structure via draft confidence, beating static or fixed-width methods.
    \item We construct small draft VLMs trained with the same data and loss as the target, improving acceptance length and reducing KL divergence via distillation.
    \item Experiments on MS COCO and out-of-domain images show up to 3.28$\times$ speedup on LLaVA-1.5 (7B/13B), with analysis of alignment, efficiency, and scalability.
\end{itemize}

\vspace{-0.5cm}

\section{Related Work}

\emph{Speculative decoding} was initially proposed to accelerate large language models (LLMs) by using a lightweight draft model to generate candidate tokens, which are then verified in parallel by a larger target model~\cite{leviathan2023fast, chen2023accelerating}. This enables lossless acceleration where outputs remain unchanged while latency is reduced. Early implementations such as Draft-and-Verify~\cite{zhang2023draft} used simple linear verification, while later approaches like Medusa~\cite{cai2024medusa} and PASS~\cite{monea2023pass} introduced multi-head decoding and parallel sampling to improve throughput. In parallel, self-executed trees and orthogonal efforts have introduced hybrid attention backbones for noise-robust perception~\cite{liu2025setransformer} and sequence-aware temporal reasoning in sports analytics~\cite{feng2025sca}. Attention-only architectures have also shown strong alignment in classical geometric estimation such as homography, which informs our visually grounded drafting and verification \cite{HuoAtten}.

Subsequent works explored tree-based speculative decoding. SpecInfer~\cite{miao2023specinfer} and EAGLE-2~\cite{li2024eagle} used static trees with fixed-width branching, which are less effective under varying draft confidence. OPT-Tree~\cite{wang2024opt} introduced adaptive branching with efficiency guarantees, while Sequoia~\cite{chen2024sequoia} applied global dynamic programming for optimal tree construction. More recent methods such as BiTA~\cite{lin2024bita} enabled lossless acceleration via bidirectional tuning and self-executed trees, and NEST~\cite{li2024nest} enhanced speculative decoding with nearest-neighbor retrieval. Hydra~\cite{ankner2024hydra} improved draft model quality through sequentially-dependent draft heads, highlighting the role of refinement.

For VLMs, prior acceleration strategies include distillation~\cite{hinton06, zhou2023distillspec}, quantization~\cite{shoeybi2019megatron}, and model simplification like MoE~\cite{rajbhandari2022deepspeed}, often trading off quality or requiring retraining. Spec-LLaVA is the first to apply speculative decoding to VLMs. Its dynamic tree-based inference with visual grounding enables lossless, efficient generation, and the lightweight draft model supports low-latency deployment in edge settings. Beyond VLM acceleration, multimodal recommendation has reported effective graph-based fusion and discrepancy reduction that improve alignment across heterogeneous modalities, including composite graph convolution with dual-stage fusion \cite{xu2025cohesion} and node–neighbor discrepancy minimization in graph convolutional networks \cite{chen2025don}.

\section{Intuition for VLM Speculation}

Speculative decoding is particularly effective for vision-language models due to several factors. First, visual inputs often provide strong grounding that constrains the space of plausible textual outputs. For example, given an image of a cat and the prompt ``What is the animal doing?'', both small and large VLMs are likely to begin with similar responses such as ``The cat is''. This visual context reduces uncertainty, increasing the likelihood that the draft model’s guesses align with the target model’s outputs. The reduced entropy in early token distributions creates favorable conditions for multi-token acceptance.

Second, many VLM tasks are descriptive or factual in nature, such as captioning or visual question answering. These outputs require less linguistic variation or creativity than open-ended text generation, making them easier for a small model to predict accurately. As a result, the draft and target distributions tend to be well aligned over many steps.

Third, we apply the same training manner to train the draft model on outputs from the target VLM. This minimizes the divergence between the two models by explicitly teaching the draft to mimic the target’s behavior, including stylistic preferences and phrasing. For example, if the target often begins answers with ``Sure, here is ...'', the draft will learn to replicate that prefix, improving acceptance. Such stylistic alignment improves not only local prefix matching but also global structural consistency.

Together, these factors contribute to long acceptance lengths during inference, even with relatively small draft models. Our empirical results confirm that VLMs are well suited to speculative decoding, achieving substantial speedup without compromising output fidelity. These properties also suggest that small, distilled draft models can serve as effective local inference agents for real-time speculative generation on resource-constrained or on-device platforms.

\section{Method}

\subsection{Draft Model Construction}

We use LLaVA-1.5 as the target vision-language model, which integrates a CLIP ViT-L/14 vision encoder with a LLaMA-based language decoder (7B or 13B parameters). To build a lightweight draft model, we construct two variants—LLaVA-68M and LLaVA-160M—sharing the same vision encoder to avoid redundant image encoding. The language decoders are significantly smaller, containing 68M and 160M parameters respectively.

The 68M model uses an 8-layer Transformer with hidden size 512 and 8 attention heads, while the 160M model employs 12 layers with hidden size 768 and 12 heads. CLIP-extracted image features are projected into the language embedding space. Both models take image-prompt pairs as input and generate speculative continuations. These compact architectures support fast speculative generation under compute and memory constraints, making them particularly suitable for deployment in edge or embedded systems.

To align the draft model with the target distribution, we apply a same training manner procedure: the draft is trained using the same multimodal instruction data as LLaVA-1.5, minimizing KL divergence with respect to the target model’s output distribution. 

\begin{figure}[h]
    \centering
    \includegraphics[width=\linewidth]{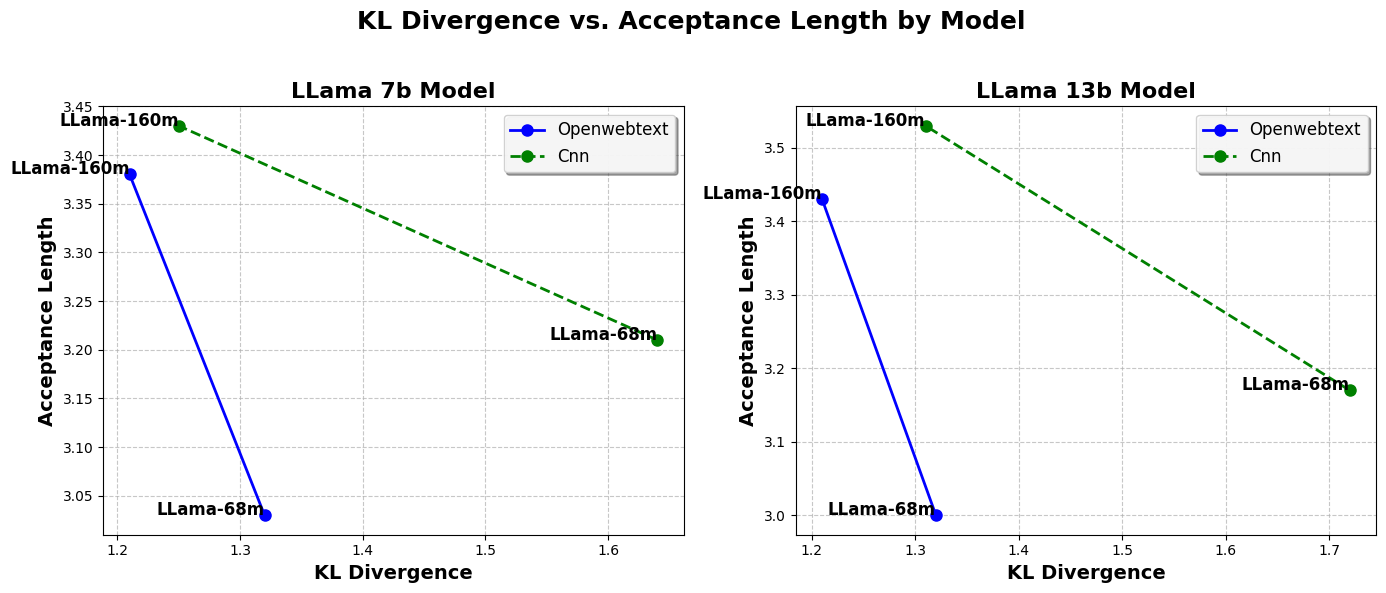}
    \caption{Lower draft-target KL divergence is associated with longer acceptance lengths, indicating better alignment.}
    \label{fig:kl-length}
\end{figure}

As shown in Fig.~\ref{fig:kl-length}, our experiments on large language models reveal a clear correlation between KL divergence and acceptance length. Specifically, a smaller KL divergence consistently leads to improved acceptance length. Motivated by this observation, we design the draft model to match the target model in both training methodology and dataset. This alignment helps maintain a low KL divergence, thereby improving the overall quality and efficiency of the decoding process.

\subsection{Dynamic Tree-Based Verification}

At inference time, the draft model generates a speculative token tree rooted at the current decoding context. The branching factor at each step is determined dynamically by the draft model’s token-level confidence: if the distribution is peaked (low entropy), the top-1 token is used; if uncertain, multiple top-$b$ tokens are expanded. During inference, the tree is pruned based on output logits, retaining only the tree structure and the top-$n$ tokens for the whole tree.

Verification proceeds in a leaf-to-root manner. The target model traverses the draft tree, comparing its predicted token at each step with the candidates generated by the draft. If a match is found at the current depth, the token is accepted and the verification proceeds to the next step. Otherwise, the speculative block is truncated, and the target model resumes greedy generation from that point onward. This conservative strategy guarantees that the final output is identical to that of the target model running alone.

Compared to static tree-based decoding (e.g., SpecInfer, EAGLE) or global dynamic programming (e.g., Sequoia), our approach performs online, heuristic tree expansion using draft model logits. It requires no offline optimization, enabling seamless integration into VLM pipelines. Inspired by OPT-Tree, we further prune invalid branches early during traversal, reducing wasted computation and enabling longer acceptance spans. This method maximizes output entropy, which increases the likelihood of accepting tokens during speculative decoding.

\section{Experiments}

We evaluate Spec-LLaVA on vision-language generation tasks to investigate speculative decoding effectiveness in multimodal contexts. Specifically, we focus on: (1) practical speedup achieved, (2) output quality preservation, and (3) the influence of draft model size and alignment on acceptance length and acceleration.

\paragraph{Setup.} We use LLaVA-1.5 (7B/13B) as target models with two lightweight drafts (68M and 160M). The evaluation includes 200 image-prompt pairs from MS COCO and a small out-of-domain set, covering descriptive captioning and visual question answering. All experiments run on a single NVIDIA L40 GPU, comparing Spec-LLaVA against baseline greedy decoding. We report wall-clock decoding times, average acceptance length ($\gamma$), and verify output exactness to baseline.

\begin{table}[ht]
\centering
\caption{Comparison of KL divergence and acceptance length for fine-tuned draft model and original model}
\begin{tabular}{llcc}
\toprule
\textbf{Target} & \textbf{Draft} & \textbf{KL} \((\downarrow\)) & \textbf{Length} \((\uparrow\)) \\
\midrule
Llama2-7B & JF68M & 1.32 & 3.03 \\
Llama2-7B & FT-JF68M & \textbf{1.19} & \textbf{3.29} \\
Llama2-13B & JF68M & 1.32 & 3.00 \\
Llama2-13B & FT-JF68M &  \textbf{1.19} & \textbf{3.27} \\
\bottomrule
\end{tabular}
\label{tab:kl_acceptance}
\end{table}

\begin{table}[t]
\centering
\caption{Acceptance length and speedup for Spec-LLaVA on LLaVA-1.5 (7B and 13B)}
\vspace{0.5em}
\begin{tabular}{llcc}
\toprule
\textbf{Target} & \textbf{Draft} & \textbf{$\boldsymbol{\gamma}$} \((\uparrow\)) & \textbf{Speedup} \((\uparrow\))\\
\midrule
LLaVA-7B & 68M  & 2.5 & 2.41$\times$ \\
LLaVA-7B & 160M & 3.5 & 3.28$\times$ \\
\addlinespace[0.2em]
LLaVA-13B & 68M  & 2.1 & 2.12$\times$ \\
LLaVA-13B & 160M & 3.0 & 2.95$\times$ \\
\bottomrule
\end{tabular}
\label{tab:main-results}
\end{table}

\paragraph{Acceptance Length and KL Divergence.} We hypothesize that reducing the KL divergence between the draft and target models leads to improved speculative decoding performance. To test this, we fine-tune the draft model on the same dataset used to train the target model, encouraging the two models’ output distributions to align more closely. As shown in Table~\ref{tab:kl_acceptance}, fine-tuning the draft model (FT-JF68M) consistently reduces the KL divergence and increases the acceptance length across both LLaMA2-7B and LLaMA2-13B target models. For instance, with LLaMA2-13B, fine-tuning reduces the KL divergence from 1.32 to 1.19 and improves the acceptance length from 3.00 to 3.27. These results validate our hypothesis and motivate our design choice: to construct the draft model for VLMs using the same training data and methodology as the target model, thereby minimizing KL divergence and enhancing speculative decoding efficiency.

\paragraph{Speedup and Acceptance Length.} Table~\ref{tab:main-results} summarizes Spec-LLaVA's performance. The 160M draft model provides up to 3.28$\times$ speedup on LLaVA-7B and 2.95$\times$ on LLaVA-13B, outperforming the 68M model due to longer acceptance spans ($\gamma$ of 3.5 vs. 2.5). Dynamic branching significantly boosts performance, particularly for smaller drafts. Allowing multiple speculative branches during uncertain predictions improves acceptance length by 15–20\% compared to a single-path strategy. In contrast, the more confident 160M draft requires fewer branches, demonstrating the adaptive efficiency of dynamic speculative decoding.



\begin{figure}[h]
    \centering
    \includegraphics[width=\linewidth]{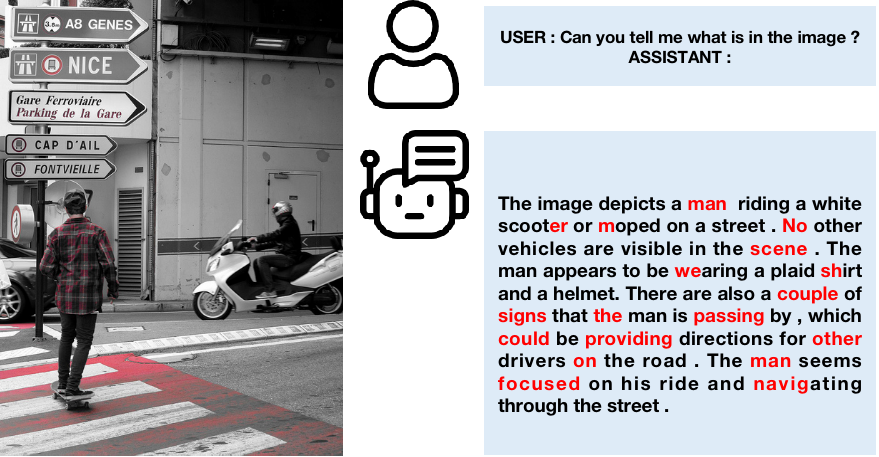}
    \caption{Speculative decoding in Spec-LLaVA (COCO example~\cite{lin2014microsoft}). Draft tokens accepted are shown vs. those verified by the target, showing efficiency.}
    \label{fig:token-acceptance}
\end{figure}

\section{Benchmarking Methodology}

To ensure reliable and reproducible speedup measurements, we benchmark all methods under controlled conditions. All models run on the same NVIDIA L40 GPU. To reduce variability from visual processing, image features are pre-extracted via the CLIP encoder. Each image-prompt pair is processed sequentially without batching to simulate realistic interactive use cases such as assistants or captioning tools. This setup also reflects low-latency scenarios typical of on-device or edge deployments.

We measure decoding latency from the first token to the final output. For speculative decoding, we log the number of verification cycles and accepted tokens per cycle. Speedup is computed as the ratio of baseline decoding time to speculative decoding time, with outputs verified as identical. All results use greedy decoding ($T=0$), ensuring determinism and removing sampling variance in timing or quality.

\section{Model Implementation Details}

We construct draft models (68M and 160M) to explore the tradeoff between size and decoding efficiency. These are empirically selected: the 68M model offers lower resource cost and faster training, while the 160M variant better aligns with the target distribution. Both reuse the CLIP ViT-L/14 encoder from LLaVA-1.5 and accept image-text inputs via LLaVA's interface. The 68M draft is especially suited for resource-constrained or on-device deployment.

Training uses 600K image-prompt pairs with AdamW (learning rate 1e-4, linear decay) for three epochs on 8 A100 GPUs with mixed precision. Distillation minimizes a weighted sum of cross-entropy and KL divergence w.r.t. target outputs. The KL term improves alignment and increases acceptance length. Checkpoints are chosen by validation acceptance length. All drafts share the target’s tokenizer; longer training improves alignment without overfitting.

\section{Conclusion}

We present Spec-LLaVA, a framework to apply speculative decoding to vision-language models in a lossless manner. By combining a compact draft model with a dynamic tree-based verification algorithm, Spec-LLaVA achieves up to 3.28$\times$ faster decoding without compromising output quality or altering outputs. Our results show that VLMs are well suited to speculative decoding due to grounded semantics, predictable output patterns, and strong alignment between visual and linguistic representations. The ability to offload draft inference to lightweight local models also makes the framework attractive for edge or on-device deployment.

This work opens several directions for future research. Combining speculative decoding with quantization or cascading draft models may yield further speedups. Extensions to multi-turn dialogues, long-form visual reasoning, and modalities like video or audio are promising. These directions could enable real-time generation for more complex multimodal systems, especially via hybrid pipelines combining local speculative generation with remote validation.

\bibliography{example_paper}

\begin{thebibliography}{40}
\providecommand{\natexlab}[1]{#1}
\providecommand{\url}[1]{\texttt{#1}}
\expandafter\ifx\csname urlstyle\endcsname\relax
  \providecommand{\doi}[1]{doi: #1}\else
  \providecommand{\doi}{doi: \begingroup \urlstyle{rm}\Url}\fi

\bibitem[Ankner et~al.(2024)Ankner, Parthasarathy, Nrusimha, Rinard, Ragan-Kelley, and Brandon]{ankner2024hydra}
Ankner, Z., Parthasarathy, R., Nrusimha, A., Rinard, C., Ragan-Kelley, J., and Brandon, W.
\newblock Hydra: Sequentially-dependent draft heads for medusa decoding.
\newblock \emph{arXiv preprint arXiv:2402.05109}, 2024.

\bibitem[Cai et~al.(2024)Cai, Li, Geng, Peng, Lee, Chen, and Dao]{cai2024medusa}
Cai, T., Li, Y., Geng, Z., Peng, H., Lee, J.~D., Chen, D., and Dao, T.
\newblock Medusa: Simple llm inference acceleration framework with multiple decoding heads.
\newblock \emph{arXiv preprint arXiv:2401.10774}, 2024.

\bibitem[Chen et~al.(2023)Chen, Borgeaud, Irving, Lespiau, Sifre, and Jumper]{chen2023accelerating}
Chen, C., Borgeaud, S., Irving, G., Lespiau, J.-B., Sifre, L., and Jumper, J.
\newblock Accelerating large language model decoding with speculative sampling.
\newblock \emph{arXiv preprint arXiv:2302.01318}, 2023.

\bibitem[Chen et~al.(2025{\natexlab{a}})Chen, Wang, Tao, Wei, Xie, Sugiyama, Raj, and Wang]{10934976}
Chen, H., Wang, Z., Tao, R., Wei, H., Xie, X., Sugiyama, M., Raj, B., and Wang, J.
\newblock Impact of noisy supervision in foundation model learning.
\newblock \emph{IEEE Transactions on Pattern Analysis and Machine Intelligence}, pp.\  1--19, 2025{\natexlab{a}}.
\newblock \doi{10.1109/TPAMI.2025.3552309}.

\bibitem[Chen et~al.(2025{\natexlab{b}})Chen, Wang, Tao, Wei, Xie, Sugiyama, Raj, and Wang]{chen2025impactnoisysupervisionfoundation}
Chen, H., Wang, Z., Tao, R., Wei, H., Xie, X., Sugiyama, M., Raj, B., and Wang, J.
\newblock Impact of noisy supervision in foundation model learning, 2025{\natexlab{b}}.
\newblock URL \url{https://arxiv.org/abs/2403.06869}.

\bibitem[Chen et~al.(2024)Chen, May, Svirschevski, Huang, Ryabinin, Jia, and Chen]{chen2024sequoia}
Chen, Z., May, A., Svirschevski, R., Huang, Y., Ryabinin, M., Jia, Z., and Chen, B.
\newblock Sequoia: Scalable, robust, and hardware-aware speculative decoding.
\newblock \emph{arXiv preprint arXiv:2402.12374}, 2024.

\bibitem[Chen et~al.(2025{\natexlab{c}})Chen, Xu, and Hu]{chen2025don}
Chen, Z., Xu, J., and Hu, H.
\newblock Don’t lose yourself: Boosting multimodal recommendation via reducing node-neighbor discrepancy in graph convolutional network.
\newblock In \emph{ICASSP 2025-2025 IEEE International Conference on Acoustics, Speech and Signal Processing (ICASSP)}, pp.\  1--5. IEEE, 2025{\natexlab{c}}.

\bibitem[Feng et~al.(2025)Feng, Ba{\v{c}}i{\'c}, and Li]{feng2025sca}
Feng, C., Ba{\v{c}}i{\'c}, B., and Li, W.
\newblock Sca-lstm: A deep learning approach to golf swing analysis and performance enhancement.
\newblock In \emph{International Conference on Neural Information Processing}, pp.\  72--86. Springer, 2025.

\bibitem[Frantar et~al.(2022)Frantar, Ashkboos, Hoefler, and Alistarh]{frantar-gptq}
Frantar, E., Ashkboos, S., Hoefler, T., and Alistarh, D.
\newblock {GPTQ}: Accurate post-training compression for generative pretrained transformers.
\newblock \emph{arXiv preprint arXiv:2210.17323}, 2022.

\bibitem[Hinton et~al.(2015)Hinton, Vinyals, and Dean]{hinton2015distilling}
Hinton, G., Vinyals, O., and Dean, J.
\newblock Distilling the knowledge in a neural network.
\newblock \emph{arXiv e-prints}, pp.\  arXiv--1503, 2015.

\bibitem[Hinton et~al.(2006)Hinton, Osindero, and Teh]{hinton06}
Hinton, G.~E., Osindero, S., and Teh, Y.~W.
\newblock A fast learning algorithm for deep belief nets.
\newblock \emph{Neural Computation}, 18:\penalty0 1527--1554, 2006.

\bibitem[Huo et~al.(2023)Huo, Ding, Xu, Tian, Zhu, Mu, Sun, Tomizuka, and Zhan]{Huo2023HumanorientedRL}
Huo, M., Ding, M., Xu, C., Tian, T., Zhu, X., Mu, Y., Sun, L., Tomizuka, M., and Zhan, W.
\newblock {Human-oriented Representation Learning for Robotic Manipulation}.
\newblock \emph{ArXiv}, abs/2310.03023, 2023.

\bibitem[Huo et~al.(2024)Huo, Zhang, Ren, Yang, and Ye]{HuoAtten}
Huo, M., Zhang, Z., Ren, X., Yang, X., and Ye, C.
\newblock {AbHE: All Attention-Based Homography Estimation}.
\newblock \emph{IEEE Transactions on Instrumentation and Measurement}, 73:\penalty0 1--11, 2024.

\bibitem[Leviathan et~al.(2023)Leviathan, Kalman, and Matias]{leviathan2023fast}
Leviathan, Y., Kalman, M., and Matias, Y.
\newblock Fast inference from transformers via speculative decoding.
\newblock In \emph{International Conference on Machine Learning}, pp.\  19274--19286. PMLR, 2023.

\bibitem[Li et~al.(2024{\natexlab{a}})Li, Zhang, Xu, Liu, Shi, and Li]{li2024nest}
Li, B., Zhang, B., Xu, Z., Liu, S., Shi, H., and Li, L.
\newblock Nearest neighbor speculative decoding.
\newblock In \emph{Proceedings of the 2024 Conference of the North American Chapter of the Association for Computational Linguistics (NAACL)}, 2024{\natexlab{a}}.

\bibitem[Li et~al.(2024{\natexlab{b}})Li, Wei, Zhang, and Zhang]{li2024eagle}
Li, Y., Wei, F., Zhang, C., and Zhang, H.
\newblock Eagle-2: Faster inference of language models with dynamic draft trees.
\newblock \emph{arXiv preprint arXiv:2406.16858}, 2024{\natexlab{b}}.

\bibitem[Lin et~al.(2024{\natexlab{a}})Lin, Yi, Li, Yang, Yu, Lu, and Xiao]{lin2024bita}
Lin, F., Yi, H., Li, H., Yang, Y., Yu, X., Lu, G., and Xiao, R.
\newblock Bita: Bi-directional tuning for lossless acceleration in large language models.
\newblock \emph{arXiv preprint arXiv:2401.12522}, 2024{\natexlab{a}}.

\bibitem[Lin et~al.(2024{\natexlab{b}})Lin, Wang, Huo, Peng, Liu, and Tomizuka]{Lin2024JointPT}
Lin, H., Wang, Y., Huo, M., Peng, C., Liu, Z., and Tomizuka, M.
\newblock {Joint Pedestrian Trajectory Prediction through Posterior Sampling}.
\newblock \emph{2024 IEEE/RSJ International Conference on Intelligent Robots and Systems (IROS)}, pp.\  5672--5679, 2024{\natexlab{b}}.

\bibitem[Lin et~al.(2014)Lin, Maire, Belongie, Hays, Perona, Ramanan, Doll{\'a}r, and Zitnick]{lin2014microsoft}
Lin, T.-Y., Maire, M., Belongie, S., Hays, J., Perona, P., Ramanan, D., Doll{\'a}r, P., and Zitnick, C.~L.
\newblock Microsoft coco: Common objects in context.
\newblock In \emph{Computer vision--ECCV 2014: 13th European conference, zurich, Switzerland, September 6-12, 2014, proceedings, part v 13}, pp.\  740--755. Springer, 2014.

\bibitem[Lin(2024)]{lin2024integrated}
Lin, Z.
\newblock \emph{The Integrated Transportation Distance Between Markov Kernels: Theory, Optimization, and Applications in Risk Evaluation and Machine Learning}.
\newblock PhD thesis, Rutgers The State University of New Jersey, Graduate School-Newark, 2024.

\bibitem[Lin \& Chen(2025)Lin and Chen]{lin2025leveraging}
Lin, Z. and Chen, Y.
\newblock Leveraging optimal transport for distributed two-sample testing: An integrated transportation distance-based framework.
\newblock \emph{arXiv e-prints}, pp.\  arXiv--2506, 2025.

\bibitem[Lin \& Ruszczynski(2023{\natexlab{a}})Lin and Ruszczynski]{lin2023fast}
Lin, Z. and Ruszczynski, A.
\newblock Fast dual subgradient optimization of the integrated transportation distance between stochastic kernels.
\newblock \emph{arXiv preprint arXiv:2312.01432}, 2023{\natexlab{a}}.

\bibitem[Lin \& Ruszczynski(2023{\natexlab{b}})Lin and Ruszczynski]{lin2023integrated}
Lin, Z. and Ruszczynski, A.
\newblock An integrated transportation distance between kernels and approximate dynamic risk evaluation in markov systems.
\newblock \emph{arXiv preprint arXiv:2311.06645}, 2023{\natexlab{b}}.

\bibitem[Lin et~al.(2025)]{lin2025federated}
Lin, Z. et~al.
\newblock Federated calculation of the free-support transportation barycenter by single-loop dual decomposition.
\newblock \emph{arXiv preprint arXiv:2507.19627}, 2025.

\bibitem[Liu et~al.(2023)Liu, Li, Wu, and Lee]{liu2023visual}
Liu, H., Li, C., Wu, Q., and Lee, Y.~J.
\newblock Visual instruction tuning.
\newblock \emph{Advances in neural information processing systems}, 36:\penalty0 34892--34916, 2023.

\bibitem[Liu et~al.(2025)Liu, Qin, Gao, Li, and Feng]{liu2025setransformer}
Liu, Y., Qin, X., Gao, Y., Li, X., and Feng, C.
\newblock Setransformer: A hybrid attention-based architecture for robust human activity recognition.
\newblock \emph{arXiv preprint arXiv:2505.19369}, 2025.

\bibitem[Miao et~al.(2023)Miao, Oliaro, Zhang, Cheng, Wang, Zhang, Wong, Zhu, Yang, Shi, et~al.]{miao2023specinfer}
Miao, X., Oliaro, G., Zhang, Z., Cheng, X., Wang, Z., Zhang, Z., Wong, R. Y.~Y., Zhu, A., Yang, L., Shi, X., et~al.
\newblock Specinfer: Accelerating generative large language model serving with tree-based speculative inference and verification.
\newblock \emph{arXiv preprint arXiv:2305.09781}, 2023.

\bibitem[Monea et~al.(2023)Monea, Joulin, and Grave]{monea2023pass}
Monea, G., Joulin, A., and Grave, E.
\newblock Pass: Parallel speculative sampling.
\newblock \emph{arXiv preprint arXiv:2311.13581}, 2023.

\bibitem[Rajbhandari et~al.(2022)]{rajbhandari2022deepspeed}
Rajbhandari, S. et~al.
\newblock Deepspeed-moe: Advancing mixture-of-experts inference and training to power next-generation ai scale.
\newblock In \emph{Proceedings of the International Conference on Machine Learning (ICML)}, 2022.
\newblock URL \url{https://proceedings.mlr.press/v162/rajbhandari22a.html}.

\bibitem[Schuster et~al.(2021)Schuster, Fisch, Jaakkola, and Barzilay]{schuster2021cats}
Schuster, T., Fisch, A., Jaakkola, T., and Barzilay, R.
\newblock Consistent accelerated inference via confident adaptive transformers.
\newblock In \emph{Proceedings of EMNLP 2021}, pp.\  4962--4979, 2021.
\newblock \doi{10.18653/v1/2021.emnlp-main.406}.

\bibitem[Shoeybi et~al.(2019)Shoeybi, Patwary, et~al.]{shoeybi2019megatron}
Shoeybi, M., Patwary, M. M.~A., et~al.
\newblock Megatron-lm: Training multi-billion parameter language models using model parallelism.
\newblock \emph{arXiv preprint arXiv:1909.08053}, 2019.
\newblock URL \url{https://arxiv.org/abs/1909.08053}.

\bibitem[Wang et~al.(2024)Wang, Su, Li, Xia, Ye, Duan, Wang, and Zhang]{wang2024opt}
Wang, J., Su, Y., Li, J., Xia, Q., Ye, Z., Duan, X., Wang, Z., and Zhang, M.
\newblock Opt-tree: Speculative decoding with adaptive draft tree structure.
\newblock \emph{arXiv preprint arXiv:2406.17276}, 2024.

\bibitem[Wang et~al.(2025)Wang, Tan, Khurana, Peri, and Ramanan]{wang2025monofusionsparseview4dreconstruction}
Wang, Z., Tan, J., Khurana, T., Peri, N., and Ramanan, D.
\newblock Monofusion: Sparse-view 4d reconstruction via monocular fusion, 2025.
\newblock URL \url{https://arxiv.org/abs/2507.23782}.

\bibitem[Wen et~al.(2024)Wen, Chen, Yan, Zhao, and Wen]{wen2024ctc}
Wen, Q., Chen, R., Yan, X., Zhao, W.~X., and Wen, J.-R.
\newblock Fast and lossless llm decoding via ctc-based drafting.
\newblock In \emph{Proceedings of NeurIPS 2024}, 2024.

\bibitem[Xu et~al.(2024)Xu, Zhao, Sun, Lin, and Han]{xu2024ondevice}
Xu, J., Zhao, Y., Sun, J., Lin, J., and Han, S.
\newblock On-device language models: A comprehensive review.
\newblock \emph{arXiv preprint arXiv:2403.05645}, 2024.

\bibitem[Xu et~al.(2025{\natexlab{a}})Xu, Chen, Wang, Hu, Kim, and Ngai]{xu2025cohesion}
Xu, J., Chen, Z., Wang, W., Hu, X., Kim, S.-W., and Ngai, E.~C.
\newblock Cohesion: Composite graph convolutional network with dual-stage fusion for multimodal recommendation.
\newblock In \emph{Proceedings of the 48th International ACM SIGIR Conference on Research and Development in Information Retrieval}, pp.\  1830--1839, 2025{\natexlab{a}}.

\bibitem[Xu et~al.(2025{\natexlab{b}})Xu, Chen, Yang, Li, Wang, and Ngai]{xu2025mentor}
Xu, J., Chen, Z., Yang, S., Li, J., Wang, H., and Ngai, E.~C.
\newblock Mentor: multi-level self-supervised learning for multimodal recommendation.
\newblock In \emph{Proceedings of the AAAI Conference on Artificial Intelligence}, volume~39, pp.\  12908--12917, 2025{\natexlab{b}}.

\bibitem[Zhang et~al.(2023)Zhang, Wang, Li, Shou, Chen, Chen, and Mehrotra]{zhang2023draft}
Zhang, J., Wang, J., Li, H., Shou, L., Chen, K., Chen, G., and Mehrotra, S.
\newblock Draft \& verify: Lossless large language model acceleration via self-speculative decoding.
\newblock \emph{arXiv preprint arXiv:2309.08168}, 2023.

\bibitem[Zhou et~al.(2023)Zhou, Lyu, Rawat, Menon, Rostamizadeh, Kumar, Kagy, and Agarwal]{zhou2023distillspec}
Zhou, Y., Lyu, K., Rawat, A.~S., Menon, A.~K., Rostamizadeh, A., Kumar, S., Kagy, J.-F., and Agarwal, R.
\newblock Distillspec: Improving speculative decoding via knowledge distillation.
\newblock \emph{arXiv preprint arXiv:2310.08461}, 2023.

\bibitem[Zhu et~al.(2023)Zhu, Tian, Xu, Huo, Zhan, Tomizuka, and Ding]{zhu2023fanuc}
Zhu, X., Tian, R., Xu, C., Huo, M., Zhan, W., Tomizuka, M., and Ding, M.
\newblock Fanuc manipulation: A dataset for learning-based manipulation with fanuc mate 200id robot, 2023.

\end{thebibliography}
\bibliographystyle{icml2025}

\newpage
\appendix
\onecolumn



\end{document}